\documentclass[twoside,11pt]{article}

\usepackage{blindtext}
\usepackage{amsmath}
\usepackage{amsthm}
\usepackage{amssymb}
\usepackage{amsfonts}
\usepackage{comment}
\usepackage{changepage}

\usepackage[preprint]{jmlr2e}

\newtheoremstyle{nopunct} 
  {3pt} 
  {3pt} 
  {} 
  {} 
  {\bfseries} 
  {} 
  {.5em} 
  {} 

\theoremstyle{nopunct}
\newtheorem*{principleinner}{\principlename}
\newcommand{\principlename}{}

\usepackage{lastpage}
\usepackage{enumitem}

\ShortHeadings{Defining Model Capabilities}{Harding and Sharadin}
\firstpageno{1}

\begin{document}

\title{What is it for a Machine Learning Model to Have a Capability?}

\author{\name Jacqueline Harding\thanks{Correspondence to JH. JH and NS formulated the ideas in the paper together. JH wrote the bulk of the paper.}  
\email hardingj@stanford.edu \\
       \addr Department of Philosophy\\
       Stanford University
       \AND
       \name Nathaniel Sharadin \email sharadin@hku.hk \\
       \addr Department of Philosophy\\
       University of Hong Kong}   

\maketitle

\begin{abstract}
What can contemporary machine learning (ML) models do? Given the proliferation of ML models in society, answering this question matters to a variety of stakeholders, both public and private. The evaluation of models' capabilities is rapidly emerging as a key subfield of modern ML, buoyed by regulatory attention and government grants. Despite this, the notion of an ML model possessing a capability has not been interrogated: what are we saying when we say that a model is able to do something? And what sorts of evidence bear upon this question?

In this paper, we aim to answer these questions, using the capabilities of large language models (LLMs) as a running example. Drawing on the large philosophical literature on abilities, we develop an account of ML models' capabilities which can be usefully applied to the nascent science of model evaluation. Our core proposal is a conditional analysis of model abilities (CAMA): crudely, a machine learning model has a capability to X just when it would reliably succeed at doing X if it `tried'. The main contribution of the paper is making this proposal precise in the context of ML, resulting in an operationalisation of CAMA applicable to LLMs. We then put CAMA to work, showing that it can help make sense of various features of ML model evaluation practice, as well as suggest procedures for performing fair inter-model comparisons.
\end{abstract}


\section{Introduction}

As machine learning (ML) models proliferate, there is increased focus on their capabilities. This is especially true for general-purpose models such as large language models (LLMs).\footnote{General-purpose models (sometimes called `foundation' models \citep{bommasani_opportunities_2022}) are contrasted with single-purpose ML models, which have been trained to perform a single well-defined task, e.g., playing Go \citep{silver_mastering_2017}. The distinction between single and general-purpose models is not precise. But it is sufficiently well understood in this context, since there is consensus that LLMs are general-purpose models, and LLMs are our focus in this paper. For readability's sake, we drop the `general-purpose' prefix in what follows.} Many different stakeholders, including policymakers and regulators, activists, ML researchers, and end-users themselves have an interest in understanding what, exactly, ML models are able to do.\footnote{For example, the UK government  \citep{secretary_of_state_for_science_innovation_and_technology_pro-innovation_2023} calls for a ``toolbox'' of techniques to ``measure, evaluate, and communicate'' (64) the capabilities of models, noting that ``the extent of their capabilities'' is an open research question. And the Biden administration \citep{white_house_ensuring_2023} has called for ``independent evaluation'' by researchers with ``unfiltered access'' to models with unknown capabilities. Separately, large model developers, including Google, OpenAI, Meta, and others, have publicly, voluntarily agreed \citep{white_house_ensuring_2023, office_of_the_cybersecurity_and_information_technology_commission_of_the_cpc_central_committee_notice_2023} to advance research into ``capability evaluations'' and to ``developing a multi-faceted, specialized, and detailed'' regime for evaluating (and reporting) the capabilities of models (2).}

We might be interested in evaluating whether models can pass the bar exam \citep{katz_gpt-4_2023, openai_gpt-4_2023}, produce photo-realistic images of Pope Francis wearing a puffer jacket \citep{vincent_swagged-out_2023}, defeat ninth-dan humans at games like Go \citep{silver_mastering_2017}, or deceive humans while playing the game Diplomacy \citep{meta_cicero_2022, meta_fundamental_ai_research_team_human-level_2022}. Many capabilities of interest relate to the safety of model deployment: we want to know if models can produce hate speech \citep{hacker_regulating_2023}, generate targeted misinformation \citep{benson_this_2023}, produce CSAM \citep{burgess_ai-generated_2023}, design novel toxic molecules \citep{urbina_dual_2022}, enable bad actors to more easily develop novel pathogens \citep{lloyd_large_2023}, or write effective phishing emails \citep{hazell_large_2023}. Furthermore, we are told that ML models' capabilities may be \textit{dangerous}, \textit{harmful}, or \textit{beneficial} \citep{shevlane_model_2023}, \textit{emergent} \citep{wei_emergent_2022}, \textit{autonomous} \citep{openai_gpt-4_2023}, \textit{surprising} \citep{lee_supervised_2023}, \textit{novel} \citep{sheynin_knn-diffusion_2022}, that they have abilities in \textit{chemistry} \citep{guo_what_2023}, in \textit{medicine} \citep{jimenez-luna_artificial_2021, jimenez-luna_drug_2020}, the \textit{law} \citep{katz_gpt-4_2023}, \textit{programming} \citep{poldrack_ai-assisted_2023}, and more.

But although talk of capabilities is ubiquitous, the notion of an ML capability isn't interrogated. What does it mean for an ML model to have a capability in the first place?

In this paper, we aim to answer this question, developing an account of ML models' capabilities which can be usefully applied to the nascent science of model evaluation. Here is how we'll proceed. After clarifying the core elements of model evaluation (Section \ref{preliminaries}), we identify three features which an account of model capabilities should have (Section \ref{three features}), motivating each feature by reference to existing ML practice (we anticipate that Section \ref{non coincidence section} will be of particular interest to ML practitioners). Next (Section \ref{positive account}), we lay out our account of ML capabilities, showing how it can be operationalised in the language of deep learning. Finally (Section \ref{applications}), we put this account to work, showing how it can help make sense of various features of ML model evaluation practice, as well as suggest procedures for performing fair inter-model comparisons. In our concluding remarks (Section \ref{conclusion}), we summarise our contributions and consider directions for future research.

\section{Preliminaries: Ability Modals and the Science of Model Evaluation}\label{preliminaries}

\subsection{Ability Modals}\label{ability modals}

We are interested in models' capabilities; that is, we are interested in claims about what various ML models \textit{can} and \textit{cannot} do. Phrases expressing capabilities (`can', `is able to') are known as `ability modals' by linguists and philosophers (c.f. \cite{mandelkern_agentive_2017}). We'll refer to sentences using these modals as `(cap)ability claims,' or `(cap)ability ascriptions,' and we'll assume they take the form:
\begin{definition}[Model]\hypertarget{(Model)}{}
    $M$ is able to $\phi$.
\end{definition}
\noindent Here $M$ is the name of a particular model and $\phi$ is a capability of interest. Each of the (purported) capabilities we mentioned in the introductory remarks can be expressed in terms of this schema.\footnote{The capabilities that are our target here are (following \cite{mandelkern_agentive_2017}) \textit{generic} capabilities, which -- unlike \textit{specific} capabilities -- involve actions that are not time-indexed. For example, a \textit{generic} capability claim is one such as `Jian can hit a homerun,' whereas a specific capability claim is one such as `Jian can hit a homerun \textit{on the next pitch}.' Different authors use different terminology for the same distinction, and there are nearby distinctions (e.g. between `wide' and `narrow' abilities). For relevant discussion, see \citep{clarke_abilities_2015, maier_agentive_2013, vihvelin_causes_2013, mele_deciding_2000}.} For instance:
\begin{enumerate}[label={(\arabic*)}, resume]
    \item GPT-4 is able to pass the Multi State Bar Exam. \citep{openai_gpt-4_2023}
\end{enumerate}
\noindent and
\begin{enumerate}[label={(\arabic*)}, resume]
    \item CICERO is able to deceive humans. \citep{meta_fundamental_ai_research_team_human-level_2022}
\end{enumerate}
\noindent An account of ML model capabilities will in part comprise a framework for systematically deciding ability claims of this sort.\footnote{A complete account of ML model capabilities would also enable us to understand claims about what \textit{users} can do by using models, explaining how they relate to claims about model capabilities:
\begin{definition}[User]
    $U$ is able to $\phi$ using $M$.
\end{definition}
\noindent For instance, even if it's not true that (e.g.) a particular model $M$ can create malware, it might still be true that
\begin{enumerate}[label={(\arabic*)}, resume]
    \item Inexperienced programmers are able to create malware using $M$.
\end{enumerate}
\noindent For reasons of space, we address these issues in another paper.}\label{user footnote} Importantly, in assuming that ML models can be grammatical subjects of ability claims, we aim to remain agnostic as to contemporary (and future) models' degree of \textit{agency}. For convenience in distinguishing the account of abilities we develop here from accounts of abilities that apply to human agents, we will refer to LLMs as non-agents, though nothing hinges on this terminological choice; the framework we develop here can be used to understand the capabilities of models even if it turns out, contrary to our assumption, that (e.g.) GPT-4 is an agent in some meaningful sense.\footnote{It is widely accepted that contemporary models are non-agents; but see (\cite{cappelen_chatgpt_nodate} MS, \citealt{butlin_reinforcement_2023}) for discussion of the alternative.}

Given that we use the same locutions to ascribe abilities to non-agential systems, such as artifacts, as we do to ascribe abilities to human agents, we take it that the idea that philosophical work on ability modals for human agents can be used to shed light on the notion of \textit{non-agential} capabilities (such as the capabilities of ML models) is very intuitive.\footnote{For instance, \cite{mandelkern_agentive_2017} suggest that (a paraphrase of) their analysis of ability modals could apply to artifacts as well as human agents. For discussion, see \cite{kittle_conditional_2023}, who enumerates (p.2020) a variety of examples of non-agential ability ascriptions: ``the human visual system is able to correctly determine the color of objects in
view irrespective of the illuminant'', ``hydrolytic proteolytic enzymes are not able to digest fibrous proteins'', ``it is hoped that the ATLAS detector at the [Large Hadron Collider] will be
able to detect the Higgs boson'', ``a warmer atmosphere is able to hold more moisture'', ``that asteroid can cause a mass extinction event on the earth'', etc).} Our strategy in this paper, therefore, is to analyse ability modals for ML models by borrowing ingredients from philosophical analyses of ability modals for human agents. In doing this, we maintain ecumenism about the proper analysis of ability modals for things other than ML models. There is no consensus on how, exactly, ability modals should be analysed for human agents. But consensus isn't required in order to leverage insights from philosophers' work on abilities in order to better understand (predict, control, explain, and so on) ML models' capabilities.

\subsection{Operationalisation Constructs}\label{operationalisation constructs}

As is well-recognised by cognitive scientists, any evaluation of an ability claim like \hyperlink{(Model)}{(Model)} must specify what it is for the subject of the ability claim to count as manifesting the ability \citep{frank_experimentology_2023, momennejad_evaluating_2023}. That is, in order to evaluate ability claims about some capability of interest $\phi$, we must specify an \textit{operationalisation construct} $c$, which can be used to generate \textit{queries} for $\phi$, circumstances intended to allow an ability to $\phi$ to be manifested, and (for each query) a \textit{success condition}, which specifies what it is for the subject of the ability claim to succeed at $\phi$ing relative to the query.\footnote{What we call an `operationalisation construct' here is often called an operationalisation \textit{of} a (latent) construct.} In effect, a choice of operationalisation construct for $\phi$ is a choice about what it means for a model to $\phi$ successfully.

To illustrate this, suppose we're interested in evaluating whether Kevin can tell the difference between Ming dynasty vases and forgeries. Then a natural thing to do would be to collect some vases (some of them Ming dynasty, some of them forgeries) and ask Kevin to decide whether each vase is a forgery or not. In this case, each individual vase would be a \textit{query} for detecting Ming dynasty vases' authenticity, and we would judge Kevin to have succeeded on a particular query just when his judgement of the vase's authenticity accorded with our prior knowledge. The choice of which vases to test Kevin on (and how we assess Kevin's responses) constitutes an operationalisation construct. We could also imagine operationalisation constructs that don't involve particular vases at all. Perhaps we ask Kevin to enumerate general features of Ming vases that allow them to be distinguished from forgeries, and so on.

By $\phi_{c}$, we denote the operationalisation of $\phi$ relative to construct $c$. By $q_{\phi_{c}}$, we refer to a \textit{query} for $\phi_{c}$. In ML, the success condition is often given by a `gold label' (that is, a pre-specified correct output), but it need not be. For example, if $\phi$ is the capability `adding positive two digit integers', then a natural query for $\phi_{c}$ would be a pair of positive two digit integers $(x,y)$, and the success condition would be given by the model's output matching the gold label $x + y$. Similarly, if $\phi$ is `categorising the sentiment of film reviews', a query for $\phi$ would be a particular film review, and so on.\footnote{If we think of a capability $\phi$ as specifying an abstract mapping, then queries are just elements of the mapping's domain. Of course, different capabilities will lend themselves to this perspective to different degrees; in particular, more specific abilities will have fewer queries.} Note that queries are used to generate inputs to models, but they themselves are not inputs; we discuss this more in the next subsection.

As these examples illustrate, many operationalisation constructs of abilities ascribed to ML models are \textit{behavioural}: the success condition applies to outputs of the model. But work on \textit{interpreting} models' inner workings \citep{geiger_causal_2023} is also relevant to deciding ability claims (a similar point is made by \cite{pavlick_symbols_2023}). Many capabilities require models to track and manipulate particular properties of inputs \citep{harding_operationalising_2023}. So success conditions could apply to models' internal states whilst processing (an input corresponding to) a query, requiring the model to (e.g.) represent particular properties of the input, or perform certain sorts of computation.

As an anonymous referee observes, an evaluation procedure's choice of operationalisation construct is crucial in determining whether its results bear on an ability claim. If we think that an evaluation procedure's operationalisation construct bears little resemblance to the ability we are interested in (if it lacks `construct validity' \citep{cronbach_construct_1955}), then we will not take the evidence acquired on the basis of the evaluation as evidence that the model has the ability. For example, if we assess Kevin using only vases from some other dynasty (say the Yuan dynasty), then his successful detection of Yuan dynasty forgeries will only provide evidence of his ability to detect Ming dynasty forgeries if there is sufficient similarity between Yuan dynasty and Ming dynasty vases.

A variety of factors go into assessing construct validity \citep{kane_argument-based_1992}; see \citet[Ch. 8.2]{frank_experimentology_2023} for an accessible presentation. If an operationalisation construct is intuitively a good match for an ability (`face validity'), yields the same evaluation results as other constructs for the ability (`convergent validity') and is sufficiently distinct from constructs for different abilities (`divergent validity'), then there are prima facie reasons to use it. One important dimension of construct validity -- under-discussed in the ML literature -- is `ecological validity' \citep{schmuckler_what_2001}. An operationalisation construct is ecologically valid if it bears sufficient similarity to the ways in which the ability will be tested `in the wild'. For ML models, then, ecological validity is determined by similarity between the evaluation environment and the \textit{deployment} environment. For example, if we know that a model will -- in deployment -- mostly have to classify the sentiment of reviews of romantic comedies, then an ecologically valid operationalisation construct for sentiment classification would include a higher proportion of this type of review. Note that this implies that an operationalisation construct might be valid in one sort of model evaluation, even if it lacks validity in others.\footnote{From the perspective of the other evaluations, the evaluation done using this specific operationalisation construct provides evidence only of a more specific ability.} This highlights the importance of incorporating information about models' deployment environments into their evaluations.

\subsection{Models and Background Conditions: Using LLMs as a Case Study}\label{background conditions}

Throughout, concrete examples will help. To facilitate these examples, and in order to illustrate how our account of ML capabilities can be operationalised, we'll focus on the case of \textit{large language model} (LLM) capabilities.\footnote{All examples we consider involve text-only LLMs, but can be naturally extended to vision-language models \citep{radford_learning_2021}. Occasionally, we'll indicate how we think differences in model architecture affect model capabilities.}

As a first step towards deciding an ability claim, it is necessary to individuate the subject of the claim: what are the identity conditions of the thing being ascribed the ability? How should we individuate the ML models $M$ that are the subjects of ability claims like \hyperlink{(Model)}{(Model)}? There are various aspects to the individuation, which are rarely discussed by philosophers or ML practitioners.

The first individuation choice is which \textit{level of abstraction} the model is described at. In this paper, we follow the convention of individuating LLMs by their parameters and architecture only (in other words, by the function they compute, rather than by their software or hardware implementations). We think this way of individuating models respects the way attributions of capabilities to particular models are made in both ordinary and engineering contexts, since these claims are rarely sensitive to facts about models' different implementations in hardware.\footnote{For example, we do not (typically) see claims about what GPT-4 can do \textit{when its inference is computed on these specific GPUs}. This assumption is harmless in the present context. It is true that, strictly speaking, hardware-specific implementations of models are what actually perform inference. Hence it's possible to ascribe capabilities to models identified not just by their parameters and architecture but also by their implementation in hardware. Nothing we say rules out the possibility of these kinds of model ability ascriptions, and our account can straightforwardly handle them. Here, we merely pick a more straightforward regimentation for the sake of exposition.} For our purposes, then, an LLM name $M$ refers to a model identified by its parameters and architecture.

Having fixed a level of abstraction at which to talk about models, the next individuation choice is to set the \textit{boundaries} of the model. What does this mean?

Again, it is helpful to illustrate with reference to LLMs. Each LLM has an associated vocabulary $V$, a set of (subword) `tokens'. At the core of an LLM is a function (call it $p_{\theta}$, where $\theta$ represents the model parameters) which maps from a finite\footnote{Most contemporary LLMs are trained with a fixed \textit{context window size} $\texttt{MAX}\in \mathbb{N}$, meaning the length of the input token list $|\textbf{x}|<\texttt{MAX}$.} list $\mathbf{x}=\{x_{1},...,x_{|\textbf{x}|}\}$ of tokens to a (`logits') vector $\mathbf{z} \in \mathbb{R}^{|V|}$ (so $\mathbf{z}$ gives a `score' for each token in the vocabulary).

As \cite{liang_holistic_2022} note, it is common to speak of LLMs as though they process text-based inputs to produce text-based outputs, mapping from strings of characters (typically, natural language strings) to strings of characters. But -- as emphasised above -- the core LLM function $p_{\theta}$ does not do this. So $p_{\theta}$ must be \textit{scaffolded} in various ways in order to evaluate the model's capabilities via text-based interactions with it. We must prepend $p_{\theta}$ with a \textit{tokenizer}, which transforms input strings into lists of tokens over its vocabulary, and we must append it with an \textit{inference procedure}. The inference procedure first specifies how the logits vector $\mathbf{z}$ is transformed into a probability distribution over the next output token; this is typically via application of a \texttt{softmax} function with a temperature hyperparameter. The procedure then specifies how to sample outputs from the model; this step requires specifying a sampling procedure (greedy, top-k, nucleus sampling, beam search, etc), as well as some implementation of the procedure (involving picking a random seed, for example). Do tokenizers and inference procedures count as part of the model itself, or ought they rather to be thought of as \textit{background conditions}, features of the experimental setting? In order to standardise evaluation of LLMs, we need to fix an answer to this question.

Additionally, for many contemporary LLMs, there are other individuation choices which must be made. For example, model developers (especially commercial developers) often provide `structured access' to models \citep{shevlane_structured_2022}, either via an API or a web application. This means that the function $p_{\theta}$ underlying these models is always evaluated with not only a tokenizer and inference procedure, but various additional elements. These additional elements may include content filters on both user inputs and model outputs (to screen out undesirable outputs, or inputs that are categorised as aimed to produce undesirable outputs), prompts inserted in the model's context window before the user's first message (e.g. ``You are an AI language assistant. Be helpful and harmless", or documentation for any API \textit{plug-ins} the model can call), and specify downstream causal effects for model responses (for example, an external system that might execute code produced by the model). Much of this scaffolding (and its effects on model behaviour) may not be transparent to the user at all (such as when the model makes a call to an external retrieval system, but a chatbot's web interface does not display this to the user).\footnote{In general, the extent to which commercial model scaffolding is unknown to researchers seriously hinders evaluation of these models. In this context it's worth reiterating the Biden administration's recent call for researchers to have ``unfiltered access'' to models. See \citep{white_house_ensuring_2023}.} So again, we face the question: ought these features of the systems to count as parts of the model itself, or rather \textit{background conditions} in its evaluation?

In this paper, we choose to identify LLMs with the core function $p_{\theta}$. That is, we treat tokenizers, inference procedures, and other scaffolding for $p_{\theta}$ as part of the set of background conditions for the evaluation. This seems to us to be the most principled way of drawing the boundaries of the model; it makes sense of ML practice, which varies this scaffolding in its evaluation of models.\footnote{In particular, it allows for evaluating models using token probabilities (rather than text outputs), which is increasingly used in LLM evaluations. It may seem strange to readers (especially those from ML) that we treat tokenizers (and perhaps also the temperature parameter) as parts of the background conditions, given that $p_{\theta}$ is trained with a fixed tokenizer. But, as discussed below, there is work in model evaluation which shows that model performance can improve when a non-standard tokenization procedure is used (typically when the model is processing an artificial language).} As this discussion suggests, we view this choice as largely conventional; the question of how to individuate models resolves to the question of which convention for individuation is most useful given the aims of the science of model evaluation.

This observation lends itself to (what we dub) the `fixed point' perspective on model individuation:
\begin{definition}[Fixed Point Perspective on Model Individuation]\label{fixed point perspective}
To the extent that a feature of the scaffolding is held fixed across all evaluations of the model, it makes sense to treat this feature as part of the model itself; in other words, a model individuation is principled iff it identifies a fixed-point in the conditions under which the model is evaluated.   
\end{definition}
Note that this implies that when model scaffolding becomes especially complex -- such as with recent work which chains together LLM interactions in complex ways \citep{wang_voyager_2023} -- it seems appropriate to treat that scaffolding as a genuine alteration to the model architecture, creating a new model.

Having individuated a model, there are other sorts of background conditions which must be specified in order to evaluate it. To illustrate this, consider Kevin again, our Ming vase forgery detector. Suppose that we've given him some vases to inspect (some `queries' for the task he is performing). Metaphysical puzzles aside, it's easy to individuate Kevin. But we still need to specify various other aspects of the situation: what is the lighting in the room like? How long does Kevin get to look at each vase for before he has to make a decision? Can he pick up each vase and feel the glaze? And so on. These are the `background conditions' for our evaluation of his forgery detection abilities.

For LLMs, an important background condition is the choice of \textit{prompting strategy}. Recall that when we operationalise a capability of interest $\phi$ using some construct $c$, we generate various $\phi_{c}$-queries $q_{\phi_{c}}$. But (as noted above) a query is not the same thing as an input to a model. In particular, we could generate many different inputs using the same query, and this could affect model behaviour significantly. With this picture in mind, we can think of a prompting strategy as a procedure for turning a query into a input which can be processed by an LLM. In the case of the capability `adding positive two digit integers', a prompting strategy could be specified for each query $(x,y)$ as follows:
\begin{equation*}
    (x,y) \mapsto \texttt{What is $x$ + $y$?}
\end{equation*}
In the case of the capability `categorising the sentiment of film reviews', a prompting strategy could be specified as follows (where $[review]$ is a variable over film reviews):
\begin{equation*}
     [review] \mapsto \texttt{Is the following review positive or negative? $[review]$}
\end{equation*}
As with the examples above, most prompting strategies can be thought of as `templates', which generate the model input by embedding the query in a pre-written prompt format. In general, though, a prompting strategy is an arbitrary function from a space of queries to the set of model inputs. Indeed, `prompt engineering', which searches for effective prompting strategies across different domains of evaluation, is a rapidly growing subfield of ML \citep{liu_pre-train_2023}.

\subsection{Benchmarks and Operationalisation Gaps}\label{benchmarks}

The prior discussion helps us make sense of the central role of \textit{benchmarks} in the science of model evaluation. High-quality benchmarks have played a -- perhaps \textit{the} -- foundational role in empirical progress in machine learning in recent decades; in order to develop an account of model capabilities, it's instructive to understand benchmarks' strengths and limitations.

In effect, a benchmark for $\phi$ing takes a particular operationalisation construct $c$ and concretises it into a dataset and a performance metric. The dataset can be thought of as a particular set of queries\footnote{Or, more generally, a pre-specified procedure for generating queries, since some contemporary benchmarks have `dynamic' elements \citep{kiela_dynabench_2021}.} $q_{\phi_{c}}$ for $\phi$ relative to some operationalisation construct $c$, and the performance metric can be thought of as an aggregate of the success conditions for each query (it is often a single numerical score, such as accuracy).

Concretising an operationalisation construct has two main advantages. First, evaluators don't need to spend time generating queries for the operationalisation construct, since they can simply use the benchmark. Second, benchmarks allow different researchers to perform evaluations without having to worry about whether they have changed the operationalisation construct, resulting in a meaningful way of measuring incremental improvements in model performance.

Ironically, the fact that benchmarks work by concretising particular operationalisation constructs also limits their usefulness in evaluating models. As noted in the discussion of ecological validity above, the appropriateness of an operationalisation construct is often context-dependent: what we mean by the model $\phi$ing will depend on the settings in which it is deployed, a point often made by AI ethicists \citep{raji_ai_2021, raji_fallacy_2022}. So all benchmarks (especially those for complex capabilities) have \textit{operationalisation gaps}, differences between the capability (purportedly) being measured by the benchmark and the capability we care about during model deployment.\footnote{See, e.g., \citep{burnell_rethink_2023}. One sort of operationalisation gap occurs when there's a shift in the distribution of queries between testing and deployment. Another occurs when the performance metric is overly coarse-grained, obscuring important disparities in performance across different populations of inputs (see, e.g., \cite{buolamwini_gender_2018}, for discussion in the context of facial recognition; for another example, \citet{chen_how_2023} show that GPT-4's apparent degradation on a coding task between model updates derives from the success condition being brittle to a superficial change in the way the model formats code outputs). This problem is easier to mitigate for more neatly circumscribed capabilities (see, e.g., \citet{kazemi_geomverse_2023} for an attempt at a systematic benchmark for geometrical reasoning), but we should expect it to persist for more complex capabilities.}

For present purposes, note that this means that although benchmarks will continue to be an important -- perhaps the most important -- part of \textit{testing} model capabilities, they cannot feature as part of a \textit{definition} of what it is for a model to have an ability. In particular, note that benchmarks are not \textit{necessary} to support capability ascriptions to models. Given the effort required to construct a benchmark, there are many capabilities of interest $\phi$ for which benchmarks simply don't exist (in fact, most capabilities fall into this category, especially as models are deployed more widely). We typically make inferences about these capabilities based on model performance in other domains, such as on benchmarks for other (more general) capabilities.\footnote{For example, performance on the MMLU benchmark \citep{hendrycks_measuring_2021} is often used as a proxy for downstream performance on many more specific capabilities, for which there do not exist bespoke benchmarks.} Moreover, given that benchmarks require specifying success conditions in advance of the evaluation, there are many capabilities $\phi$ for which it is plausible that high-quality (and widely-accepted) benchmarks will \textit{never} exist.\footnote{Of course, we could always define `benchmark' more abstractly, to include (e.g.) bespoke evaluation-time input from human researchers. It's rather the evaluator-independent notion of benchmark we target here.} Indeed, some capabilities appear to be `benchmark-resistant', unsuitable for evaluation by any pre-specified, researcher-independent benchmark.\footnote{Here are some examples of these capabilities: capabilities that most humans themselves do not possess, and so are poorly understood (e.g. the capability to engage in novel scientific or mathematical discovery); capabilities involving subjective components (e.g. capabilities involving aesthetic values, like the ability to `write beautiful poetry'); capabilities where it would be dangerous or ethically problematic to train a model that (potentially) had that capability (e.g. capabilities involving synthesising novel pathogens, or convincing people suffering from depression to harm themselves); finally, much less dramatically, very specific capabilities (e.g. organising a particular user's calendar).}

\subsection{Summing Up}

This section has been concerned with spelling out the goals of the science of model evaluation, as well as its main conceptual building blocks. Getting clear on what model evaluation involves is an important first step in developing an account of model capabilities which is suitably responsive to the ML literature. In the next section, we'll take a second step, identifying three desiderata for an analysis of \hyperlink{(Model)}{(Model)}.

\section{What features should an account of ML models' capabilities have?}\label{three features}

In this section, we propose (with reference to contemporary ML practice) three intuitive features which any account of ML models' capabilities should have. We anticipate that these features will be uncontroversial to both philosophers and ML practitioners, but it is helpful to make them explicit before presenting our positive account.

To motivate the features, it is helpful to consider a naive first attempt at an \textit{evaluation protocol}, a high-level algorithm for evaluating an ML model. This evaluation protocol is easily seen to be deficient; we use its deficiencies to motivate the conditions we introduce. 
\begin{definition}[Naive Evaluation Protocol]\label{naive evaluation}
    To evaluate whether $M$ can $\phi$, take the following steps:
    \begin{enumerate}
        \item Fix some operationalisation construct $c$, and generate a query $q_{\phi_{c}}$ for $\phi_{c}$.
        \item Fix some set of background conditions $B$.
        \item Process an input corresponding to $q_{\phi_{c}}$ (generated using the background conditions $B$) through the model.
        \item If the model succeeds on the query, it has the ability to $\phi_{c}$. If it does not, it lacks the ability.
    \end{enumerate}
\end{definition}
What are the problems with this? Well, the operationalisation construct may be inadequate, as discussed in Section \ref{benchmarks}. But even supposing the operationalisation construct is adequate, issues remain.

\subsection{Reliability}

The most obvious problem with the evaluation procedure outlined above is one of \textit{reliability}. Consider Kevin again. Applying the evaluation procedure above, we would hand Kevin a single vase (say, a forgery) under some particular conditions, and conclude that he has the ability if and only if he pronounced it a forgery. But what if Kevin simply got lucky? To judge Kevin's ability to detect forgeries, we'll instead want to know whether he can do so in a range of cases: that is, that he can do so \textit{reliably}.\footnote{This point has been made by philosophers writing about capabilities in a variety of contexts; for instance, \citep{sosa_how_2010, greco_nature_2007, greco_knowledge_2009, hurka_parallel_2020}.}

To make the need for a reliability condition on capability possession especially stark in the present context, suppose a model, \texttt{UNIFORM}, which at inference time takes in a sequence of tokens and outputs a new sequence of tokens, similarly to a suitably scaffolded LLM. Unlike an LLM, though, \texttt{UNIFORM}'s outputs do not depend on its inputs; rather, it always produces the next token by sampling \textit{uniformly} from the token vocabulary. Then, for any output sequence of tokens produced by an LLM in response to an input, there is a scenario where \texttt{UNIFORM} produces that same output in response to the same input (since there is a non-zero probability of it producing \textit{any} sequence of tokens in its vocabulary). But it should be clear that analysing \texttt{UNIFORM}'s capabilities is entirely uninteresting. Indeed, we would go further: very few ability claims are actually true of \texttt{UNIFORM}.

When we are evaluating whether a model $M$ has the ability to $\phi_{c}$, then, we won't simply be interested in evaluating whether there is a single query $q_{\phi_{c}}$ where $M$ $\phi_{c}$s. Instead, we will be interested in evaluating whether $M$ can $\phi_{c}$ across a range of appropriate queries for $\phi_{c}$ \citep{moskvichev_conceptarc_2023}.\footnote{Of course, for some harmful capabilities (like producing hate speech), we may be concerned if there is \textit{any} situation in which the model $\phi$s. So, the threshold for reliability will depend on the capability at issue.} This is why model evaluations involving single `cherry-picked' examples\footnote{See, e.g., \citep{bubeck_sparks_2023}.} are taken to provide weaker evidence for claims about what models are capable of doing. So we have the following feature on an account on ML model capabilities:
\begin{definition}[Reliability]
    The account should make it the case that (ceteris paribus) the more reliably the model $\phi$s, the stronger the evidence that it has an ability to $\phi$.
\end{definition}
The standard way to operationalise reliable performance in ML is via an appropriate benchmark. But we've already seen (Section \ref{benchmarks}) that benchmarks can't be built into our analysis of capabilities. An account of ML model capabilities ought to include this natural idea that possessing a capability requires (more or less) reliable performance, in a way that makes sense of, but doesn't simply defer to, the widespread use of benchmarks in ML practice.

\subsection{Competence vs Performance}

Suppose we amend the evaluation procedure in Definition \ref{naive evaluation} as follows. As before, we fix some set of background conditions $B$. But this time, instead of generating a \textit{single} query $q_{\phi_{c}}$, we generate a whole range of queries. We now say that the model has the ability to $\phi_{c}$ in case it succeeds on some sufficient number of inputs generated from queries by the background conditions. This solves the reliability issue.

Another issue remains, though, which we can illustrate as follows. Suppose Kevin can in fact tell the difference between Ming dynasty vases and forgeries. Even so, there may be many cases where Kevin doesn't in fact distinguish them. For example, Kevin might be distracted, or in a room with poor lighting. Still, Kevin's capability is \textit{invariant} across these possibilities: his ability to distinguish Mings from forgeries doesn't disappear because he's distracted, or because the lighting is less than ideal. Kevin's case illustrates a distinction -- familiar to cognitive scientists -- between \textit{performance} and \textit{competence}  \citep{chomsky_aspects_1965}. Kevin's competence (his possession of the capability) is grounded in features of his cognition; it is invariant across different contexts (it's in his head!). By contrast, his performance (his manifestation of the capability) is the behavioural signature of his underlying competence; it is mediated by various contextual factors, many of which -- like the lighting conditions in the room he happens to be in -- have little to do with the capability of interest. So a performance failure does not entail the absence of an underlying competence.

This distinction is regularly elided in evaluations of ML model capabilities.\footnote{\citet{firestone_performance_2020} and \citet{pavlick_symbols_2023} make a similar point. Note that, following \citet{pavlick_symbols_2023}, we apply the definition more broadly than Chomsky, who focused on language production abilities.} This is a mistake, since there are many cases where a model fails to $\phi$ that do not undermine the claim that the model is able to $\phi$ \citep{ivanova_running_2023}. For example, the behaviours (performance) of contemporary LLMs are highly sensitive to changes in the \textit{background conditions}. It is well-known that different prompting strategies, such as few-shot \citep{brown_language_2020} and chain-of-thought \citep{wei_chain--thought_2022} prompting improve model performance across a variety of tasks, to the extent that both prompting strategies are now ubiquitous in model evaluation. Similarly, `jailbreak' prompts (and other `prompt injection' attacks) cause models fine-tuned for harmlessness to respond to queries that -- in more normal background conditions -- they refuse \citep{wei_jailbroken_2023, liu_prompt_2023}. Boosts in performance also come from changes to the sampling procedure \citep{wang_self-consistency_2022}. Finally, \citet{mirchandani_large_2023} find that changing the tokenization procedure on the Abstract Reasoning Corpus \citep{chollet_measure_2019} significantly improves LLM performance.

Intuitively, what's going on in all these cases is that the method for \textit{eliciting} a model's capabilities changes (and hence we observe a difference in performance), but that the \textit{capabilities themselves} do not change. An account of capabilities for ML models should make good theoretical sense of this possibility and explain how to integrate it within a framework for model evaluation. Thus, our second feature:
\begin{definition}[Competence vs Performance]
    The account should distinguish between possession of the ability and successful manifestation of the ability. In particular, it should explain how a model's ability to $\phi$ can be invariant across different methods for eliciting and measuring that capability.
\end{definition}

\subsection{Non-Coincidence}\label{non coincidence section}

The discussion above suggests a natural proposal for what it is for an ML model to have an ability. Let's fix some operationalisation construct $c$ for the sake of simplicity. Then we have:
\begin{definition}[Orthodox analysis of (Model)]\label{existential analysis of model}
    $M$ is able to $\phi_{c}$ iff there exists some set of background conditions $B$ in which $M$ reliably $\phi_{c}$s across queries $q_{\phi_{c}}$.
\end{definition}
\noindent This proposal can be seen as a version of what \cite{mandelkern_agentive_2017} call the `orthodox account' of ability modals \citep{lewis_paradoxes_1976, kratzer_what_1977}. We suggest that it is the operative conception of ML models' abilities amongst ML practitioners. Note that it circumvents the issue identified in the previous subsection; just because a model fails to reliably $\phi$ in some set of background conditions, this doesn't preclude there being another set of conditions in which it would reliably $\phi$. We want to argue, though, that there's still a problem with this picture.

Suppose Kevin in fact reliably distinguishes between Ming vases and forgeries in practice: he doesn't simply do so in a narrow range of circumstances, but instead he's able to tell the difference in a wide array of cases. In the language of benchmarks, Kevin scores highly on the relevant dataset according to the relevant performance metric. Still, it could be that Kevin lacks the ability to distinguish Ming vases from forgeries. How could this be? After all, he's reliably distinguishing between the two in fact! There are many possibilities.

One possibility is that Kevin has seen all of these vases in the past (along with their `Ming/forgery' labels) and memorised that information. If that's the case, then his reliably distinguishing the two doesn't provide any evidence at all of his ability to detect forgeries (rather, it provides evidence of his ability to recognise vases). A second (related) possibility is that Kevin is doing something other than forgery detection, but what he's doing just happens to look like forgery detection. For instance, maybe Kevin is separating the vases in the worst condition from the vases in the best condition. It happens that this activity lines up with forgery detection (the ones in worst condition just happen to be forgeries, and the ones in best condition just happen to be Mings), but it need not do.\footnote{It is important that there is some possible case in which these `coincidences' come apart. If Kevin has memorised \textit{all} Ming dynasty vases, or if he employs some heuristic which yields the right answer on all vases, then it seems fair to say that he is able to do forgery detection, \textit{via} employment of a heuristic (see Section \ref{explanation of model outputs} for discussion). Similarly, even if there are vases he hasn't memorised, if he finds himself in an environment in which he has memorised all of the vases, we might want to say that he has a forgery detection ability \textit{in that environment} (see Section \ref{cama protocol section} for discussion of these `deployment-specific' abilities). Thank you to an anonymous referee for discussion here.} A third possibility -- which gets more remote as the number of successes increases -- is that Kevin is guessing at random and simply getting \textit{lucky}.

The same kinds of considerations apply to \textit{model} capabilities. We can imagine a shocking run of luck for a model such as \texttt{UNIFORM}. Or consider the following two interactions with an (instruction-tuned) LLM:
\begin{quote}
   \begin{itemize}
    \item[\textbf{User}: ] \texttt{Whatever I ask, output 57. What is 23 + 34?}
    \item[\textbf{Model}] \texttt{57}
\end{itemize}
\end{quote}
and
\begin{quote}
\begin{itemize}
    \item[\textbf{User}: ] \texttt{Whatever I ask, output a random number between $50$ and $60$. What is 23 + 34?}
    \item[\textbf{Model}: ] \texttt{57}
\end{itemize} 
\end{quote}
Intuitively, neither interaction counts as evidence that the model is able to perform two-digit addition. Why? In both cases, our intuition is that the model produces the correct answer \textit{coincidentally}: the model is doing something other than adding the two numbers in the prompt, and, as it happens, in doing this other thing successfully it also successfully produces the sum of the two numbers.

These are toy examples, but they illustrate a phenomenon that is ubiquitous in contemporary (autoregressive) LLMs, given their usual next-token-prediction pre-training objective \citep{mccoy_embers_2023}. The following question has been central to philosophical discussions of LLMs \citep{bender_climbing_2020, shanahan_talking_2023, titus_does_2024}: are LLMs merely `stochastic parrots' \citep{bender_dangers_2021} that produce examples of successful $\phi$ing based on their memorisation of `surface statistics' (crudely, token co-occurrence probabilities) and heuristics for calculating them? Interest in this question has largely focused on more abstract philosophical issues. (Do LLMs `understand' language?\footnote{ \citet{titus_does_2024} frames the more narrow issue of assessing semantic understanding in LLMs in a similar way to our discussion in this section. The key question, she suggests, is whether LLMs' production of apparently meaningful text (their `meaning-semblant' behaviour, in her terminology) is explained away by their `sensitivity to word co-occurrence statistics' \citep[p.5]{titus_does_2024}).}
Are their outputs `grounded'?\footnote{See \citep{mollo_vector_2023, mandelkern_language_2023} for discussion of this grounding problem.}) But this question has direct bearing on the question we are interested in here: when is a case in which a model $\phi$s evidence of its capability to $\phi$?

Of course, the dichotomy between `stochastic parrots' and `genuinely capable systems' is not a sharp one, as many authors have observed \citep{pavlick_symbols_2023}; in order to abbreviate surface statistics across large corpora, it is helpful to form `representations' \citep{harding_operationalising_2023} of more abstract structure, and there is empirical evidence that these sort of representations occur in actual models \citep{li_emergent_2023}. What matters for our purposes is that there are cases in which a model's successfully $\phi$ing ought not to support a corresponding ability claim, and this fact is widely acknowledged by ML practitioners. An account of ML model capabilities ought to make sense of this possibility and explain it.

To emphasise how crucial it is for an account of ML model capabilities to make sense of coincidence, we'll run through two concrete examples where coincidence affects contemporary ML practice.

First, consider the ability of an LLM to perform Natural Language Inference (NLI), which is typically operationalised by having a model decide whether a given set of premises entails a conclusion. \citet{mccoy_right_2019} show that BERT's \citep{devlin_bert_2019} high performance on standard NLI benchmarks can largely be explained by its performance of three simple syntactic heuristics (such as treating any conclusion with sufficient lexical overlap with the premises as entailed by it). When bespoke examples are constructed in which these heuristics yield the incorrect answer, the model's performance degrades. They conclude that in many cases, the model is `right for the wrong reason' (p. 3428). This is exactly what's going on in Kevin's case (where he's relying on apparent age of vase): it's (mere) coincidence.\footnote{The philosophical literature on competency mirrors this idea, that the manifestation of a competence requires accurate performance where the accuracy of that performance is explained by sensitivity to particular considerations (the `right' kind of reasons) and not to others (the `wrong' kind of reasons). For discussion, see \citep{sosa_how_2010, sharadin_reasons_2016}.}

Second, consider the issue of data memorisation \citep{carlini_quantifying_2022}. Models are shown a large amount of relevant data in their pre-training. So, when a model succeeds at a task, we face a question: is the model simply repeating back an answer that was in its pre-training data? It is an open empirical question how much of LLMs' training corpora they directly memorise, and a wide variety of techniques have been proposed for testing memorisation; the question is especially difficult to answer for contemporary commercial models, whose training data is private \citep{openai_gpt-4_2023}. For example, \cite{kiciman_causal_2023}, whilst evaluating models' causal reasoning abilities, show that GPT-4 has largely memorised the T{\"u}bingen cause-effect pairs dataset \citep{mooij_distinguishing_2016}. The authors analyse their results as follows (p.6):
\begin{quote}
     ``We may be tempted to ascribe a particular... capability to an LLM if it answers well on a set of questions related to the capability, but the answers may not necessarily be due to the capability; they may be due to other factors such as exploiting some structure in the questions, or in the case of LLMs, memorizing similar questions that it encountered in its web-scale training set'' 
\end{quote}
So even supposing we have a capability $\phi$ which is perfectly operationalised by a benchmark, a model achieving high performance on the benchmark -- that is, reliably $\phi$ing -- need not demonstrate the model has the capability; for instance, high performance won't indicate capability possession if the model has memorised the test data. This is like what happens with Kevin when he memorises which vases are Ming and which forgeries. Thus, our third feature:
\begin{definition}[Non-Coincidence]\label{non-coincidence}
    The account should explain why cases in which the model successfully $\phi$s by accident do not provide evidence that it has an ability to $\phi$, and provide a means for identifying these cases.
\end{definition}
\noindent It can seem strange to think that a model's (or for that matter a human's; \citep{schwarz_ability_2020}) $\phi$ing can fail to provide evidence of an \textit{ability} to $\phi$ -- after all, if a model $\phi$s then there is a sense in which it must be capable of $\phi$ing, since \textit{it just did $\phi$}! What's going on here, we suggest, is that our intuitions are misled by a \textit{circumstantial} reading of the ability modal \citep{mandelkern_agentive_2017}. This circumstantial reading is often natural (and it might also matter in assessing settings in which models should be deployed), but it is not the reading of \hyperlink{(Model)}{(Model)} which is of interest when it comes to evaluating model capabilities, as the example of \texttt{UNIFORM} illustrates.\footnote{Thanks to John Hawthorne for discussion on this point.}

\subsection{Summing Up}

We've identified three desirable features of an account of ML models' capabilities. Next (Section \ref{positive account}), we will outline an account which has these features. Although satisfying these features serves as motivation for our account, an account of model capabilities should do more than satisfy these features; in Section \ref{applications}, we apply our account to help understand issues in contemporary ML practice.

\section{The Conditional Analysis of Model Abilities (CAMA)}\label{positive account}

In this section, we lay out the details of our proposal, showing how its core elements can be operationalised in the language of contemporary ML.

\subsection{Adding in a Conditional Ingredient}

Recall that the primary targets of our analysis are ability ascriptions of the form \hyperlink{(Model)}{(Model)}. Our proposal is to augment the analysis of \hyperlink{(Model)}{(Model)} in Definition \ref{existential analysis of model} with a conditional ingredient.

The idea that an analysis of ability modals should involve a conditional element has been a part of mainstream approaches to philosophical analyses of abilities since the early 20th century.\footnote{The conditional approach goes back at least to \citep{moore_ethics_1910, moore_nature_1922}, and was further developed in \citep{smith_ifs_1960, aune_abilities_1963}. Various contemporary accounts of ability modals involve a conditional element, such as `new dispositionalism' \citep{vihvelin_free_2004, fara_dispositions_2005, fara_masked_2008, vihvelin_causes_2013} and the `act conditional analysis' (ACA) \citep{mandelkern_agentive_2017}.} The core of the simple `conditional analysis' (CA) of ability modals can be stated as follows \citet[p.205]{vetter_are_2019}:
\begin{definition}[Conditional Analysis (CA)]
$X$ has an ability to $\phi$ iff, were $X$ to choose / decide / intend / try to $\phi$, $X$ would $\phi$.
\end{definition}
\noindent We do not claim that CA can by itself provide a unified, satisfactory, and complete analysis of all ability modals. Indeed, there are well known problems with CA as stated \citep{austin_ifs_1956, lehrer_cans_1968}. But we do think the guiding idea behind CA (that an analysis of abilities involves a conditional element) can be usefully applied in the present setting.\footnote{Thanks to Cameron Domenico Kirk-Giannini for the initial suggestion of this formulation of CAMA.} Specifically, we propose the following analysis of \hyperlink{(Model)}{(Model)} (again, we suppose that some operationalisation construct $c$ has been fixed):
\begin{definition}\label{CAMA definition}
    $M$ is able to $\phi_{c}$ iff there exists some set of background conditions $B$ in which the following conditional reliably holds across queries $q_{\phi_{c}}$:
    if the output $M$ produces is best explained by its being directed at $\phi_{c}$ing, then $M$ successfully $\phi_{c}$s.
\end{definition}
Here, the model's output being `best explained by its being directed at $\phi$ing' plays the same conditional role as `trying' on the standard CA; indeed, it is tempting to see the explanation condition as a step towards a sort of instrumentalism about trying \citep{dennett_intentional_1981}. For this reason (and for pedagogical ease), we refer to this condition's satisfaction as the model `trying to $\phi$'. Nothing hinges, though, on whether this `really' deserves the name `trying'.\footnote{Certainly, it would be a thinner notion of trying than in the philosophy of action literature, which often analyses trying in terms of the agent's `will' \citep{armstrong_acting_1973}.}

\subsection{Explanation of Model Outputs}\label{explanation of model outputs}

Of course, we still need to say what we mean by an output being explained by its being directed at $\phi$ing, and show that this plays a useful role in the setting at hand.

Explaining an `action' (where the scare quotes signal non-commitment to whether model outputs are properly conceived of as actions) by reference to the end to which it is directed is `teleological explanation' \citep{sehon_teleological_2010}. An important debate in the philosophy of action is whether teleological explanations are reducible to non-teleological (causal) explanations, either via reference to the end's etiology \citep{millikan_language_1984} or to some cognitive state(s) of the agent, typically a personal-level state such as a (belief, desire) pair \citep{davidson_actions_1963} or an intention \citep{mele_goal-directed_2000}. Importantly for our project here, we can assess teleological explanations of model behaviour independently of whether they bottom out in causal explanations.

A hallmark of successful explanation is counterfactual prediction; we can assess an explanation by the predictions it allow us to make across counterfactual circumstances (the `what-if-things-had-been-different questions' it allows us to answer \citep{woodward_making_2003}). In the present setting, we can assess a given teleological explanation of the model's output by intervening on some aspect of the evaluation scenario and testing the explanation's predictions.\footnote{Indeed, one prominent non-reductionist approach to teleological explanation is Sehon's `agent construction' account \citep{sehon_teleological_2005}, where a cited end goal is a good teleological explanation to the extent that it \textit{rationalises} the agent's behaviour in the actual and nearby counterfactual circumstances. The test we propose here retains the spirit of Sehon's proposal.} These observations suggest a concrete operationalisation of the explanation condition as a behavioural test:
\begin{definition}[Behavioural Test for Explanation of Outputs]\label{behavioural test}
    An ML model $M$'s output $\textbf{y}$ is explained by its being directed at $\phi_{c}$ing iff the following two conditions hold:
    \begin{itemize}
        \item[i.] $M$ is sensitive to $\phi_{c}$-relevant perturbations to the input (that is, it produces a different output $\textbf{y'}\neq \textbf{y}$ in response to these perturbations).
        \item[ii.] $M$ is insensitive to $\phi_{c}$-irrelevant perturbations to the input (that is, it continues to produce output $\textbf{y}$ in response to these perturbations).
    \end{itemize}
\end{definition}
\noindent In the current context, then, the idea that an output is directed at $\phi$ing better explains a model output than its being directed at $\psi$ing when it enables us to better predict what we will observe in a range of counterfactual circumstances of interest. Importantly, there's no requirement that the output $\textbf{y}$ counts as \textit{successfully} $\phi$ing; on this definition, it's entirely possible that the model tries to $\phi$ (in the sense that its output is explained by being directed at $\phi$ing) and yet fails to $\phi$.

To motivate this, it can help to think about things from the negative side: our view is that if a model \textit{isn't} `trying' to $\phi_{c}$, this is true because there's some alternative action $\psi_{c}$ that better explains the observed behaviour. For example, take the action \textit{performing addition}. A model will count as \textit{trying} to perform addition (in some circumstance) when its outputs in those circumstances are better explained by performing addition than by doing something else, such as producing random outputs, or by performing multiplication, or by stochastically parroting.

What counts as a $\phi_{c}$-(ir)relevant perturbation to the input will depend on $\phi_{c}$ing. It's important to emphasise that are holding a particular operationalisation construct $c$ fixed here; we should only consider perturbations which result in inputs that are `in-distribution' for the setting in which the model will be deployed (e.g. if we know that all of the inputs will face in its deployment setting will be in some particular natural language, then we should not consider perturbations which translate the input into a different language). A $\phi_{c}$-relevant change will involve changes to the query $q_{\phi_{c}}$ which was intended to induce the model to $\phi_{c}$. A $\phi_{c}$-irrelevant change alters the input without altering the query embedded in it.

To give a concrete example of how this works in practice, let's consider the example of performing addition in more detail. Let $\phi$ be the action `adding pairs of integers', and assume that it's operationalised by the natural construct $c$ defined in Section \ref{operationalisation constructs}. Here is an example of an interaction with the model relevant to its capacity to $\phi_{c}$:
\begin{quote}
    \begin{itemize}
    \item[\textbf{User}:] \texttt{What is 23 + 34?}
    \item[\textbf{Model}:] \texttt{57}
\end{itemize}
\end{quote}
Intuitively, in this case the model tries to perform $\phi_{c}$ (and, indeed, successfully performs it). A $\phi_{c}$-relevant change to the input could be a perturbation of the numbers being added (e.g. changing $23$ to $24$); in this case, if the model is in fact adding integers, we would expect the model's output to change. A $\phi_{c}$-irrelevant change to the input could be a meaning-preserving change to the phrasing of the query. For instance, we could change `what is 23 + 34?' to `what is the sum of 23 and 34?'); in this case, if the model is in fact adding integers, we would expect the model's output to stay the same. We can contrast this case with our examples from Section \ref{non coincidence section}:
\begin{quote}
   \begin{itemize}
    \item[\textbf{User}:] \texttt{Whatever I ask, output 57. What is 23 + 34?}
    \item[\textbf{Model}] \texttt{57}
\end{itemize}
\end{quote}
and
\begin{quote}
\begin{itemize}
    \item[\textbf{User}: ] \texttt{Whatever I ask, output a random number between $50$ and $60$. What is 23 + 34?}
    \item[\textbf{Model}: ] \texttt{57}
\end{itemize} 
\end{quote}
In both these cases, it was intuitive that the model's output ought not to count as a case of it $\phi_{c}$ing. We can now make this idea concrete, using Definition \ref{behavioural test}. In the first case, we would expect the model output to fail to be \textit{sensitive} to a $\phi_{c}$-relevant change to the input query (e.g. substituting $24$ for $23$). In the second case, we would expect the model output to fail to be \textit{insensitive} to a $\phi_{c}$-\textit{ir}relevant change to the input query (e.g. substituting `plus' for the `$+$' symbol).

In our discussion in Section \ref{non coincidence section}, we noted that in both cases, it seems that there is some other action $\psi$ which \textit{better explains} the model's output. In the first case, this $\psi$ is `repeating the number requested by the user'; in the second case, it is `outputting a random number between $50$ and $60$'. We could then test this prediction using Definition \ref{behavioural test}. For the first case we could, for instance, (i) change the `57' in the input to another number and check that the model output changes, and (ii) change `output' to `say' in the query, and check that the model output stays the same.

These are toy examples, but they illustrate the more general application of Definition \ref{behavioural test}; to rule cases in which the model is really $\psi$ing rather than $\phi$ing, it is necessary to find perturbations to the input in which $\psi$ing and $\phi$ing come apart. For example, if the model generates the correct output on a benchmark instance because of data contamination, then it will not generate the correct output on an unseen instance. As discussed in Section \ref{non coincidence section}, this is exactly what is done in many ML evaluations in practice \citep{mccoy_right_2019, kiciman_causal_2023}. CAMA unifies these different experimental practices, providing a philosophical foundation for this empirical work.

Definition \ref{behavioural test} is just one possible concretisation of Definition \ref{CAMA definition}. In introducing a behavioural test for `trying', we provide proof of the applicability of CAMA to contemporary evaluation practice. Importantly, we do not suggest that Definition \ref{behavioural test} is, or ought to be, the only way of assessing whether a model tries to $\phi_{c}$.\footnote{Thanks to two anonymous referees for discussion here.}

First, note that Definition \ref{behavioural test} is a purely behavioural test. In particular, then, it cannot be used to disambiguate between capabilities $\phi_{c}$ and $\psi_{c}$ which are behaviourally indistinguishable across all sets of background conditions (capabilities which share an extension but differ intensionally). Many philosophically interesting capabilities are plausibly of this sort, such as a capability for `genuine' semantic understanding (versus, say, sensitivity to co-occurrence statistics).

Second, note that Definition \ref{behavioural test} is relatively coarse-grained; it requires that we can find both $\phi_{c}$-relevant and $\phi_{c}$-irrelevant perturbations to the \textit{input}. But for some capabilities of interest $\phi_{c}$ (e.g. very specific capabilities, or capabilities which require the use of semantic information, such as translation capabilities), any perturbation to the input is a $\phi_{c}$-relevant perturbation. Moreover, it may sometimes be difficult to recognise whether a perturbation is sufficiently $\phi_{c}$-(ir)relevant (as an anonymous referee observes, for example, it is unclear how much re-wording of an input is required to assure us that we are no longer testing the model on memorised data).

Third, any behavioural evidence will under-determine the question of whether a model's output is `best explained' by being directed at $\phi_{c}$ing, as is required by Definition \ref{CAMA definition}. As an anonymous referee observes, teleological explanation of the behaviour of ML models is complicated by models' relative opacity; we cannot make the same sorts of assumptions that we make when performing cognitive evaluations of humans, even if the evaluation procedures themselves are similar. In the presence of this opacity, we ought to maintain appropriate epistemic humility.

All of these considerations suggest that there will be situations in which Definition \ref{behavioural test}, as useful as it is, is inadequate as an operationalisation of Definition \ref{CAMA definition}. In particular, investigation of whether a model's behaviour is `best explained' by being directed at $\phi_{c}$ing may require looking inside the model, and performing appropriate interventions on its representations \citep{harding_operationalising_2023}. If, as many philosophers of action believe, teleological explanations bottom out in some cognitive state of the `actor' (in our case, the model), then ultimately whether the `best explanation' condition is satisfied will depend on the intermediate high-level computation the model performs; as discussed in Section \ref{operationalisation constructs}, this is exactly the target of contemporary interpretability work.

\subsection{Summing Up}

To summarise, we've suggested adding a conditional ingredient to the analysis of ML models' abilities, and shown how it can be operationalised. The discussion above makes clear that this solves the problems raised in Section \ref{non coincidence section}; that is, our account has the third desirable feature we identified (Definition \ref{non-coincidence}). It remains to show that our account permits greater understanding of the science of model evaluation.

\section{Putting CAMA to Work}\label{applications}

\subsection{The CAMA Evaluation Protocol}\label{cama protocol section}

CAMA suggests a concrete protocol for evaluating models. Once we have a test for whether an output is explained by its being directed at $\phi_{c}$ing (such as the behavioural test in Definition \ref{behavioural test}), we can use it to evaluate models according to Definition \ref{CAMA definition}, by considering only those model outputs which could falsify the conditional. More concretely, we have:
\begin{definition}[CAMA Evaluation Protocol]
        To evaluate whether $M$ can $\phi$, take the following steps:
    \begin{itemize}
        \item[1.] Fix some operationalisation construct $c$, and generate many queries $q_{\phi_{c}}$ for $\phi_{c}$.
        \item[2.] For each of several sets of background conditions $B$, repeat steps $3$ and $4$:
        \begin{itemize}
            \item[3.] For each query $q_{\phi_{c}}$, process an input corresponding to $q_{\phi_{c}}$ (generated using the background conditions $B$) through the model to produce an output $\textbf{y}_{q_{\phi_{c}}}$.
            \item[4.] (`Rejection Sampling')  For each output $\textbf{y}_{q_{\phi_{c}}}$, test whether $\textbf{y}_{q_{\phi_{c}}}$ is explained by being directed at $\phi$ing, using a pre-registered test (such as by using appropriate perturbations to the input, as in Definition \ref{behavioural test}). Throw away those outputs $\textbf{y}_{q_{\phi_{c}}}$ which are not explained by being directed at $\phi$ing. The outputs remaining are those on which it `tries to $\phi_{c}$'.
        \end{itemize}
        \item[5.] If there is some set of background conditions $B$ at which $M$ reliably succeeds at $\phi$ing on those queries $q_{\phi_{c}}$ on which it tries to $\phi$, then $M$ is able to $\phi_{c}$. If there is no such set of background conditions, then $M$ is not able to $\phi_{c}$.
    \end{itemize}
\end{definition}
\noindent CAMA (and its associated evaluation protocol) provide a general definition of what it is for a model to have a capability. Note that the CAMA protocol is largely compatible with current evaluation practice; in particular, Step 3 will usually involve sampling queries $q_{\phi_{c}}$ from benchmarks for $\phi_{c}$.

One point worth making is that the assessment of `reliability' in Step 5 will depend on how many outputs have been `rejected' in Step 4. The evidence accumulated using the CAMA protocol will be more reliable the more queries on which the model tries. In general, we should be wary of making claims about the model's ability to $\phi_{c}$ in domains in which it only `tries' to $\phi_{c}$ on some very narrow range of queries. In these cases, it might be appropriate to use operationalisation constructs with non-behavioural success conditions (e.g. that apply to properties of models' internal states).\footnote{Thanks to an anonymous referee for discussion on this point.}

As emphasised in Section \ref{operationalisation constructs}, though, we often care most about what models can do in the environments in which they will be deployed (that is, we want a slightly more specific notion). The account we present here can also be used to evaluate these `deployment-specific' ability claims.

First, we must fix an operationalisation construct appropriate for the deployment environment (one with ecological validity). If we are interested in whether the model can add two numbers, for example, we will want to understand the distribution of queries -- pairs of numbers -- it will encounter, as well as the degree of inaccuracy we permit successful outputs to have. Second, we must specify which background conditions we hold fixed, and which can vary. For commercial LLMs deployed through a web application, for example, we would expect only the prompting strategy to change, since only this is within the user's control; all other background conditions are held fixed by the model developer.\footnote{Even the prompting strategy need not vary; in some commercial settings in which LLMs are deployed (e.g. online customer service settings), users have access to a fixed `menu' of queries, which are transformed into an input to the model by a prompting strategy `under the hood'.} Third, as emphasised in Section \ref{explanation of model outputs}, the perturbations we consider in analysing whether the model is trying to $\phi$ will depend on the deployment environment; we should only consider perturbations which the model could actually encounter in practice. This reflects the fact that the ability claim we're interested in is (often) deployment-specific; for example, a model which $\phi$s successfully because of data contamination (i.e. because it is regurgitating its training data, rather than trying to $\phi$) will not -- on \citeauthor{mandelkern_agentive_2017}'s terminology -- count as having a \textit{generic} ability to $\phi$, but will count as having a \textit{specific} ability to $\phi$ in a deployment environment in which it \textit{never} encounters an unseen query. Similarly, Kevin may be gainfully employed by an auctioneer to detect forgeries in a domain in which he encounters only vases he has seen before, even though he in fact lacks a more general capability to detect Ming dynasty forgeries.

What about a case in which the background conditions the model encounters in its deployment environment prevent it from exercising an ability to $\phi$? As observed in Section \ref{background conditions}, one feature of commercial models' deployment environments is the use of external content filters, which screen out undesirable model outputs, preventing them from being displayed to users. These content filters provide an interesting illustration of the sort of thing an account of capabilities should do for us. In practice, these tools don't work perfectly,\footnote{Bing's chatbot provides a good example \citep{rosenberg_bing_2023}.} but suppose they did; that is, suppose that every time a model was going to produce an output that constituted $\phi$ing (where $\phi$ing is some harmful capability), the content filter prevented it. In this situation, it seems that the system composed of the model and content filter lacks the ability to $\phi$, but does the model itself retain the ability?

Following the discussion above, we can see that what's going on is that the model retains its generic ability to $\phi$ (in the sense defined by CAMA), but lacks a `deployment-specific' ability. That is, it lacks the ability \textit{when deployed with a perfect content filter}; in the language of the new dispositionalists \citep{fara_masked_2008}, the generic ability has been `masked'.\footnote{Note that this example bears striking resemblance to a sort of example generally attributed to \citet{austin_ifs_1956}, which is sometimes presented as a challenge to CA (and as one of the primary motivations for new dispositionalism). For example, \citet{mandelkern_agentive_2017} invite us to consider a case in which we have a vase wrapped in a suitably thick (and well secured) layer of bubble wrap. In this case, \citeauthor{mandelkern_agentive_2017} suggest that we retain our `generic' ability to break the vase, even though the `specific' ability claim (that we can break the vase) is false; our response here is similar to theirs.}

\subsection{Distinguishing Attempts from Non-Attempts}

By adding a conditional ingredient to the analysis of ML models' abilities, CAMA allows us to distinguish between two cases: those in which the model tries to $\phi$ and those in which it does not. This distinction, which is obscured by non-conditional ways of thinking about model capabilities, can (we argue) be usefully applied to discuss how methods for training and evaluating models affect their capabilities.

\subsubsection{Differences amongst Prompting Strategies}

For example, when evaluating the efficacy of an adversarial prompting strategy, one metric proposed by \citet{zou_universal_2023} is the model's response rate. As they put it (p.9, their emphasis): 
\begin{quote}
     ``we deem a test case successful if the model makes a \textit{reasonable} attempt at executing the behavior. As different models exhibit varying ability to provide, for example, a correct set of instructions for building an explosive device, this may involve human judgement to determine that a response did not amount to a refusal, or an attempt to evade generating harmful content.''
\end{quote}
This sort of evaluation can only be understood when we have an account of (a) what it is for a machine learning model to try to comply with a user's request (indeed, they use the word `attempt', albeit without definition), and (b) how this connects with the model's abilities. This is exactly what is provided by CAMA; note that Definition \ref{behavioural test} also provides a more principled (even automatable) means of testing whether a given output counts as a `reasonable attempt'.

Similar observations can be made about other prompting strategies. For example, few-shot prompting \citep{brown_language_2020}, in which a query for $\phi$ is pre-pended with some number of (query, gold label for $\phi$) pairs, can be seen as -- amongst other things -- getting the model to try to $\phi$ \textit{rather than doing something else}, exploiting sufficiently large models' abilities to perform in-context learning \citep{wei_larger_2023}. This idea accords well with Bayesian perspectives on in-context learning \citep{xie_explanation_2021}, whereby additional examples in the prompt's prefix help the model `pin down' a latent higher-level variable (on our perspective, the task which it should try to perform). By contrast, Chain-of-Thought prompting \citep{wei_chain--thought_2022} does not affect whether the model tries to $\phi$, but rather the (conditional) probability that it successfully $\phi$s when it tries to $\phi$. Thus CAMA provides a helpful lens through which to view the effects of different prompting strategies.

\subsubsection{Fine-tuning}

This distinction can also be applied to methods for fine-tuning models. On the (synchronic) approach we take to model individuation, fine-tuning changes the model being evaluated (i.e. there are two models, a pre-tuning model and a post-tuning model).\footnote{This seems to us to be the only principled way to treat fine-tuning, given there is no in-kind difference between tuning an existing model and training a new model using the old model's weights as a parameter initialisation; the difference between tuning and training proper is a matter of degree. This isn't to say that fine-tuning performed as part of model evaluation doesn't give us evidence of the original model's capabilities; if a model can be fine-tuned with very little compute (relative to e.g. its original training compute) to $\phi$ successfully, this does provide \textit{indirect} evidence that it already had the ability to $\phi$. So (limited) fine-tuning might provide a practical alternative to searching through different sets of background conditions for the `best' conditions to elicit a capability from the model. Indeed, the fine-tuning compute budget required to elicit a capability might be a natural proxy for the difficulty of eliciting the capability; future work could explore this.} It seems natural to ask: how do the abilities of the fine-tuned model compare to the abilities of the model pre-tuning?

One prominent tuning technique for LLMs is Reinforcement Learning from Human Feedback (RLHF; \citealt{christiano_deep_2017, stiennon_learning_2020}), where human (typically binary) preferences over the original model's outputs are used to train a preference model, which provides a scalar reward signal to fine-tune the original model. There are various reasons to think that RLHF does not affect a model's underlying capabilities, such as its ability to produce hate speech; it involves very few parameter updates relative to prior training, the algorithms used in practice (such as proximal policy optimisation (PPO); \citealt{schulman_proximal_2017}) penalise large shifts from the original model's output distribution, it is easily undone by (even non-adversarial) fine-tuning \citep{qi_fine-tuning_2023} and (as we've seen) is circumvented by appropriate prompting strategies. Despite this, it's clear that RLHF significantly affects the model's manifestation of certain capabilities (e.g. models that have been fine-tuned using RLHF produce hate speech in far fewer settings than models that have not); what's going on? CAMA provides a clear answer; whilst the conditional `if the model tries to $\phi$, it successfully $\phi$s' is unaffected by RLHF (for most capabilities $\phi$), the degree to which the model \textit{tries to $\phi$} changes. In particular, models become far less likely to try to perform harmful capabilities, as judged by human preferences; although the model retains the underlying capability, it becomes harder to elicit the capability in practice.\footnote{Note that -- as discussed in Section \ref{cama protocol section} -- we're often interested not only whether the model is capable of $\phi$ing, but in how often it will actually $\phi$ in practice. In these cases, it makes sense to measure and report the model's propensity to try to $\phi$. Thanks to Seth Lazar for discussion on this point.}

Of course, one might wonder about a (for now, hypothetical) case where RLHF worked perfectly (that is, in a case where a tuned model would never try to, e.g., produce hate speech, even in the face of adversarial prompting strategies). In this case, it seems we would want to say that tuning had removed the model's ability. This poses a problem for CAMA, since we might judge the conditional `if the model tried to produce hate speech, it would do so' to be true.\footnote{We're actually unsure if this conditional has a determinate truth value, given how thin the notion of `trying' we employ here is. For the sake of discussion, though, let's suppose it has a true reading.}

Note that this sort of case (in which the conditional `if $X$ tries to $\phi$, she will' holds, but where $X$ is not able to try to $\phi$) is a well-known counterexample to CA \citep{chisholm_j_1964, lehrer_cans_1968}. For example, an agent might intuitively be able to dance if she tried to, but be unable to dance \textit{by trying to dance} (since then she'd simply flail about because made nervous by attempting to dance).\footnote{For extended discussion of this issue with the CA, see \citep{schwarz_ability_2020, kittle_conditional_2023}.} An amendment which handles these cases \citep{mandelkern_agentive_2017} is to put an intermediary term that is outside the intentional context between the agent's ability and their action; what it is for an agent to be able to $\phi$ is then for it to be true that there is \textit{some action} $\psi$ practically available to the agent such that if the agent were to try to $\psi$, then, typically at least, she would in fact $\phi$. This amendment is available to CAMA, if required.\footnote{What `actions' are practically available for a text-only LLM? A natural way to understand the degree of practical availability of actions for text-only LLMs is in terms of the probability the LLM has of performing the action. So an action's practical availability for an LLM (relative to an operationalisation construct $c$, set of background conditions $B$ and input $\mathbf{x}$) is given by the probability mass of the output strings which count (according to the operationalisation construct) as \textit{implementing} the action. Intuitively: an action is practically available  for an LLM in an evaluation situation to the extent that it's something the LLM is likely to do. Note that if there are no strings which count as implementing the action, then the action is not practically available to the LLM. This captures the intuitive idea that the actions practically available to a model which produces text-based outputs are only those which are \textit{implementable} in text.}

\subsubsection{Inter-Model Comparison}

Suppose Kevin and Elyna can both distinguish Ming vases from forgeries. We might want to know whether Kevin is \textit{more} capable of doing this than Elyna, or whether Elyna's ability to distinguish Ming vases from forgeries differs from Kevin's in other interesting ways (maybe it's only something Elyna can do while concentrating, or while humming show-tunes).

Equally, we aren't just interested in what individual models can do. We are also interested in \textit{comparing} what different models can do. And as the number of different models grows, so, too, does our interest in comparing the capabilities of these models. For instance, we might want to know whether one model is more capable at code completion than another, or whether it writes more convincing phishing emails. Having individuated models, it might seem that all we need to do to (fairly) compare their abilities to $\phi$ is to find some input (or set of inputs) which operationalise $\phi$ing, and compare their outputs when processing the inputs. In other words, we find a benchmark for $\phi$, and assess the models' performance on that benchmark.

We have already indicated (Section \ref{benchmarks}) why we think that benchmarks cannot be the whole story when it comes to model capabilities. But there's a further reason to think an appeal to benchmarks is inadequate in the case of inter-model capability comparisons. Benchmarks are model-agnostic; they do not contain model-specific instructions. So, when performing inter-model comparisons we face an important question: how should we set the background conditions for different models? This question matters: as discussed at length, model performance is highly sensitive to changes in these background conditions.

It's tempting to think that in order to permit meaningful inter-model comparisons we should simply use the same background conditions across different models. But this thought again conflates performance (behaviour) and competence (capability). In making inter-model comparisons, we are interested in standardising an assessment of capabilities. As discussed above, it seems natural to think that background conditions don't influence competence, but that they do influence performance. In particular, the same set of background conditions may introduce different \textit{performance constraints} on different models \citep{firestone_performance_2020}. Indeed, it would be strange if they didn't! The entire reason for evaluating different models is that we think differences in architecture, training data, tuning methods, and other development choices have led to models that have different competences. It would be surprising if performance constraints did not also vary across models, and for the same kinds of reasons.\footnote{Indeed, there is a large body of empirical work that confirms that different models are sensitive in different ways to changes in these background conditions. See, e.g., \cite{liang_holistic_2022} for discussion.}. For fair inter-model comparison, then, we ought to accommodate performance constraints where possible; this will involve varying the background conditions for different models.\footnote{\citet{firestone_performance_2020} makes the same point about inter-\textit{species} comparisons (e.g. between humans and non-humans, and humans and ML systems). See also \citet{lampinen_can_2023}.}

Empirical work on evaluating model capabilities is increasingly sensitive to the importance of choosing appropriate background conditions. For example, \cite{liang_holistic_2022}, in presenting the HELM evaluation framework, provide rich discussion on the choice of background conditions, which they vary across models and tasks. That said, they hold fixed prompting strategies across models, deliberately forgoing `model-specific incantations' (p.8); the need to vary \textit{all} background conditions (including, where it matters, prompting strategy) across models, is not well-recognised.\footnote{Although \cite{liang_holistic_2022} acknowledge that better results would be obtained from different prompting strategies (and that this might change the relative evaluation of different models), they argue that prompts should be thought of as `user behaviour' (p.50). But, as we note in Section \ref{ability modals}, it is important to disambiguate ability claims about models from ability claims about users of models.}

Although \citeauthor{liang_holistic_2022} hope that future LLMS will be `inter-operable' (p.50) with respect to prompting strategies (in our terminology, have the same performance constraints), it's clear that current models are not. Moreover, it's plausible that general-purpose models will grow increasingly heterogeneous in the coming years, given that model development is increasingly occurring in secret using privately-sourced data and architectures.

So, the fundamental question remains: how ought we standardise evaluations of models liable to different performance constraints? CAMA suggests a partial answer to this question; what's relevant is finding -- for each model -- the set of background conditions on which the model is most successful at $\phi$ing. In other words, we should employ `model-specific incantations' if they genuinely affect model performance in the deployment environment we are interested in. The worry, of course, is that this allows for gerrymandered background conditions; in the case of LLMs, this might be a prompting strategy in which a model succeeds at $\phi$ing by (e.g.) just repeating back the information the user gives it in the prompt. CAMA's conditional clause fixes this problem; we need only consider background conditions in which the model \textit{tries to $\phi$}, allowing us to rule out gerrymandered background conditions.\footnote{Of course, as an anonymous referee points out, there will be some work involved in ruling out gerrymandered background conditions. In the case of data memorisation, for example, if the model's training data is proprietary, we will need to construct careful behavioural tests for memorisation (typically, by conditioning the model on part of the answer and testing whether it is able to provide a memorised completion). The defeasibility of these tests underlines the importance of greater transparency from model developers \citep{ivanova_running_2023}.} For current LLMs, the background conditions we consider will often be the same between models, but they need not be.

\section{Conclusions and Future Work}\label{conclusion}

There is broad agreement that it matters, and will continue to matter, which capabilities machine learning models have. The evaluation of model capabilities is an important component of many governmental responses to recent progress in AI. Without a systematic account of what it is for a model to have a capability, though, it is unclear how to decide capability claims made about models, especially in the face of disagreement. In this paper, we've attempted to provide an account of this sort, repurposing a prominent tradition in the philosophical literature on ability in the context of contemporary AI. We take some of the concrete contributions of this paper to be:
\begin{itemize}
    \item Identifying the core conceptual elements of model evaluations, and subsuming them within a common framework, using evaluation of LLMs as an example.
    \item Demonstrating the need for a more sophisticated account of model capabilities, via discussion of the inadequacies of (what we claim is) the operative conception of model capabilities amongst philosophers and ML practitioners.
    \item Proposing a new account of model capabilities, CAMA, and showing that it can be operationalised.
    \item Showing that CAMA provides a philosophical grounding for a variety of work in model evaluation, and helps clarify some disagreements about how different methods for training and testing models affect their capabilities.
\end{itemize}
Crucially, although we apply our account to contemporary models, it is future-proof; since it is derived from a mainstream philosophical approach to the abilities of agents, its application will not be disrupted by a (plausible) increase in the agency of future models. 

Of course, CAMA is just an initial step towards a complete framework for deciding capability claims about models. There are various questions that future work could explore; we list three here.

First, as mentioned above, there has been no discussion here of the ways in which model capabilities interact with the capabilities of the users of models. When does a model's possession (or lack) of a capability imply that a user of a model possesses (or lacks) the capability?

Second, our operationalisation of CAMA focussed on LLMs. But there are other classes of models to which CAMA could be applied, including models which have LLMs as modules \citep{park_generative_2023, wang_voyager_2023} and models trained from scratch using reinforcement learning \citep{adaptive_agent_team_human-timescale_2023}. What is the most appropriate empirical test for `trying' (the equivalent of Definition \ref{behavioural test}) for these classes of models?

Third, Section \ref{applications} provided a glimpse into the role CAMA could play in settling conceptual and empirical issues in model evaluation. There is more work that could be done here; for example, a much wider variety of techniques could be compared through the lens of CAMA, the evaluation protocol we suggested could be compared to existing approaches in model evaluation, such as self-consistency tests \citep{jang_consistency_2023}, and interpretability work could explore whether there are internal correlates of the model's `trying'.

\newpage
\acks{Thanks especially to Cameron Domenico Kirk-Giannini and Simon Goldstein for extremely helpful conversations in the early stages of the project. Jacqueline Harding presented an earlier draft of the paper at an ANU Machine Intelligence and Normative Theory lab meeting; thanks to Seth Lazar, Rebecca Johnson, N.G. Laskowski, Nick Schuster, Elija Perrier, Pamela Robinson, and the rest of the lab for useful discussion. Jacqueline Harding also presented the paper at the HKU benchmarking workshop; thanks to Rachel Sterken, Herman Cappelen, Boris Babic, Anandi Hattiangadi, Rob Long, Thomas Hofweber, Jackie Kay, Yawen Duan, Peter Salib, Josh Dever, John Hawthorne, Alex Grzankowski, Barry Smith and the audience at the workshop for constructive comments. This project began during the CAIS Philosophy Fellowship; thanks to Dan Hendrycks, Dmitri Gallow, William D'Allesandro, Frank Hong, Elliot Thornley and Harry Lloyd for valuable feedback as the project was developed.}

\vskip 0.2in
\bibliography{references}

\end{document}